\documentclass[10pt,twocolumn,letterpaper]{article}

\usepackage[pagenumbers]{cvpr} % To force page numbers, e.g. for an arXiv version

\usepackage{graphicx}
\usepackage{amsmath}
\usepackage{adjustbox}
\usepackage{amssymb}
\usepackage{booktabs}
\usepackage{verbatim}
\usepackage{tablefootnote}
\usepackage{stfloats}

\usepackage[normalem]{ulem}
\usepackage{ifthen}
\usepackage{enumitem}
\usepackage{multicol}
\usepackage{multirow}
\newcommand{\anna}[2][]{%
    \ifthenelse{ \equal{#1}{} }
        {\textcolor{blue}{(Anna) #2}}
        {\textcolor{blue}{(Anna) \sout{#1\xspace}#2}}
}
\newcommand{\George}[2][]{%
    \ifthenelse{ \equal{#1}{} }
        {\textcolor{magenta}{(George) #2}}
        {\textcolor{magenta}{(George) \sout{#1\xspace}#2}}
}

\newcommand{\todo}[2][]{%
    \ifthenelse{ \equal{#1}{} }
        {\textcolor{red}{TODO: #2}}
        {\textcolor{red}{TODO: \sout{#1\xspace}#2}}
}

\newcommand{\cat}[0]{2,091 }
\newcommand{\dog}[0]{3,274 }

\usepackage[pagebackref,breaklinks,colorlinks]{hyperref}

\usepackage[capitalize]{cleveref}
\crefname{section}{Sec.}{Secs.}
\Crefname{section}{Section}{Sections}
\Crefname{table}{Table}{Tables}
\crefname{table}{Tab.}{Tabs.}

\def\cvprPaperID{*****} % *** Enter the CVPR Paper ID here
\def\confName{CVPR}
\def\confYear{2024}

\begin{document}

\title{DogFLW: Dog Facial Landmarks in the Wild Dataset}

\author{George Martvel$^{*1}$ \and Greta Abele$^2$ \and Annika Bremhorst$^2$ \and Chiara Canori$^3$ \and Nareed Farhat$^1$ \and Giulia Pedretti$^3$ \and Ilan Shimshoni$^1$ \and Anna Zamansky$^1$}
\date{} 

\maketitle

\noindent
\begin{flushleft}
$^1$ University of Haifa, Israel\\
$^2$ Dogs and Science, Switzerland\\
$^3$ University of Parma, Italy
\end{flushleft}

\bigbreak

%%%%%%%%% ABSTRACT
\begin{abstract}

Affective computing for animals is a rapidly expanding research area that is going deeper than automated movement tracking to address animal internal states, like pain and emotions. Facial expressions can serve to communicate information about these states in mammals. However, unlike human-related studies, there is a significant shortage of datasets that would enable the automated analysis of animal facial expressions.
Inspired by the recently introduced Cat Facial Landmarks in the Wild dataset, presenting cat faces annotated with 48  facial anatomy-based landmarks, in this paper, we develop an analogous dataset containing \dog annotated images of dogs. Our dataset is based on a scheme of 46 facial anatomy-based landmarks. 

The DogFLW dataset is available from the corresponding author upon a reasonable request.
%The dataset is available at \url{https://gitlab.com/is-annazam/dogflw}.

\end{abstract}

\begin{comment}
Similarly to humans, facial expressions in animals are closely linked with emotional states. However, in contrast to the human domain, automated recognition 
of emotional states from facial expressions in animals is underexplored, mainly due to difficulties in data collection and establishment of ground truth concerning emotional states of non-verbal users. 
We apply recent deep learning techniques to classify (positive) anticipation and (negative) frustration of dogs on a dataset collected in a controlled experimental setting.
We explore the suitability of different backbones (e.g. ResNet, ViT) under different supervisions to this task, and find that features of a self-supervised pretrained ViT (DINO-ViT) are superior to the other alternatives.
To the best of our knowledge, this work is the first to address the task of automatic classification of canine emotions on data acquired in a controlled experiment. 
\end{comment}

\section{Introduction}
\label{sec:intro}

Due to the internal nature of affective states, including emotions and pain, their recognition in animals is very challenging in light of the lack of a verbal basis for communication. Nevertheless, observing subtle changes in their facial expressions and body language shows promise as a non-intrusive method for studying these states. Mammals produce facial expressions, which have been linked emotional states in a variety of species~\cite{descovich2017facial}. Therefore, it is unsurprising that the number of works addressing automation of animal affect recognition tasks is rapidly growing; Broome et al.~\cite{broome2023going} provides a comprehensive review of state-of-the-art works using computer vision to address such recognition in animals. Despite dogs being one of the most well-studied animals in behavior studies, only a few of these works address these species, with solely one work~\cite{boneh2022explainable} focusing on facial expressions and using a dataset collected in a controlled experimental setting. 

The interest in cognitive and behavioral aspects of dogs has been increasing dramatically~\cite{maclean2021new}. Dogs can serve as valuable clinical models for numerous human disorders, owing to their large size: many canine conditions mirror human diseases, including diabetes, cancers, epilepsy, eye diseases, autoimmune diseases, and other rare diseases~\cite{hytonen2016canine}. Other factors contributing to the widespread interest in dogs include fascination with their origins in the context of domestication, as well as their behavior and cognitive abilities. Moreover, we must improve our understanding of and regulate dog-human interactions and welfare impacts, including working dogs and shelter dogs~\cite{maclean2021new}.

The objective measurement of dog facial expressions is crucial for their investigation as indicators of emotional states in different contexts~\cite{bremhorst2019differences, pedretti2022audience}. Facial expressions have also been studied in the context of understanding dog-human communication (e.g., the impact of dog facial phenotypes on their communication abilities with humans \cite{sexton2023written}, the impact of facial features on the ability of humans to understand dogs \cite{eretova2024can}, etc.)

The gold standard for objectively assessing changes in facial expressions in human emotion research is the Facial Action Coding System~--- FACS \cite{Ekman1978}. FACS has recently been adapted for different non-human species, including dogs~\cite{waller2013paedomorphic}. DogFACS has been applied in several studies~\cite{caeiro2017dogs, bremhorst2019differences, bremhorst2021evaluating, pedretti2022audience, sexton2023written} to objectively measure facial changes. However, using this method for facial expression analysis depends on laborious manual annotation, which also requires extensive human training and certification and may be prone to human error or bias~\cite{hamm2011automated}. Some first steps to automated DogFACS were taken by Boneh et al.~\cite{boneh2022explainable}.

Geometric morphometrics offers an appealing alternative approach, successfully applied in analyzing cat facial features~\cite{finka2019geometric, finka2020application}. This method uses points (landmarks) on objects as proxies for shape, allowing for the quantification of facial shape changes. This concept is closely related to landmarks extensively studied in the human domain.

Indeed, numerous fundamental methods for detecting, labeling, and aligning facial and body landmarks in humans have emerged~\cite{bincon, pip, he16resnet, mobilenet, graph, pose2016, PLFD, anchorface}. 
Typically, datasets with human facial landmarks consist of thousands of images with dozens of landmarks. This abundance of data leads to better model performance, even in challenging scenarios such as occlusions or low-quality images. 

The animal domain, on the other hand, severely needs landmark-related datasets. Table \ref{datas} shows the available datasets, including their size and number of landmarks, which generally are extremely small compared to the human domain.

\begin{table}[t!]
  \centering
  \begin{tabular}{@{}lllc@{}}
    \toprule
    Dataset & Animal & Size & Landmarks \\
    \midrule
    Khan et al. \cite{AW} & Various & 21,900 & 9\\
    Zhang et al. \cite{original_dataset} & Cat & 10,000 & 9 \\
    Liu et al. \cite{liu2012dog} & Dog & 8,351 & 8\\
    Cao et al. \cite{cao2019cross} & Various & 5,517 & 5\\
    Mougeot et al. \cite{mougeot19dogs} & Dog & 3,148 & 3 \\
    Martvel et al. \cite{martvel2023catflw} & Cat & \cat & 48\\
    Sun et al. \cite{sun20cafm} & Cat & 1,706 & 15\\
    Pessanha et al. \cite{pessanha2022facial} & Horse (tilted) & 952 & 44 \\
    Hewitt et al. \cite{hewitt2019pose} & Sheep & 850 & 25\\
    Yang et al. \cite{yang15sheep} & Sheep & 600 & 8\\
    Pessanha et al. \cite{pessanha2022facial} & Horse (frontal) & 370 & 54 \\
    Pessanha et al. \cite{pessanha2022facial} & Horse (side) & 348 & 45 \\
    \midrule
    DogFLW & Dog & \dog & 46\\
    \bottomrule
  \end{tabular}
  \caption{Comparison of animal facial landmarks datasets}
  \label{datas}
\end{table}

In the case of domesticated species, which are often artificially selected for specific features, there is a great variation in facial morphology and appearance. It is particularly noticeable in the case of cats and even more so in dogs, making it challenging to create versatile landmark detection models~\cite{caeiro2017development, correia2021extending, waller2013paedomorphic}. With the increase in quantity and quality of animal facial landmark datasets, it has become possible to create cross-view models~\cite{transferlandmarks2018, cao2019cross, interspecies2017}, but the number of landmarks in such works is usually relatively small. To solve the problems of determining the emotions or states of animals, several dozens of landmarks are needed~\cite{feighel, ferres2022predicting, evangelista2019facial}, which are unique for a particular animal. 

To tackle the shortage of datasets, the Cat Facial Landmarks in the Wild (CatFLW) dataset was recently introduced by Martvel et al.~\cite{martvel2023catflw}. The dataset comprises over 2,000 images of cat faces in various environments, each with face bounding boxes and 48 facial landmarks as detailed in~\cite{finka2019geometric}. Notably, the utilized landmark scheme is based on the cat's facial anatomy and the CatFACS annotation system~\cite{caeiro2017development}. This approach has facilitated the development of models for automated identification of pain in cats~\cite{martvel2023automatedA,feighelstein2023explainable, martvel2023automated}.

Dogs exhibit a wide range of facial features, making them more challenging than cats for automated facial analysis. To fill this gap, this paper introduces the Dog Facial Landmarks in the Wild (DogFLW) dataset, comprising a set of 46 facial landmarks that were defined, based on and guided by the facial anatomy of dogs and the DogFACS method. We utilized the Ensemble Landmark Detector (ELD)~\cite{martvel2024automated} to provide a benchmark for this dataset.

\section{Landmark Scheme}

\begin{figure}[b!]

\includegraphics[width=0.47\textwidth]{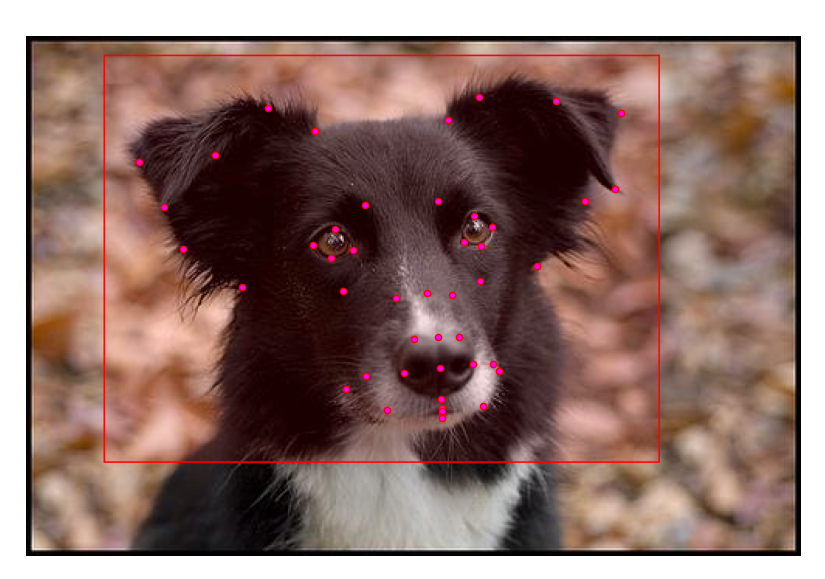}
\caption{\textbf{Annotated Dog's Face.} Image of a dog with a face bounding box and 46 facial landmarks.}\label{ann_cat}\vspace{-0.25em}
\end{figure}

The development of the dog facial landmark scheme was undertaken with the aim of creating a comprehensive framework for analyzing canine facial expressions with the help of three experienced dog behavior researchers certified in DogFACS coding\cite{waller2013paedomorphic}. To establish the number of landmarks and each landmark location, the experts worked independently and then converged using expert consensus. The obtained final landmark scheme is shown in Figure \ref{ann_cat}. All details and landmark descriptions are available in the \href{https://docs.google.com/spreadsheets/d/1M0fFfigULMu-qup-PjmE6_n_tzAJcZDiSGxMrCo6_ck/edit?usp=sharing}{spreadsheet}.

\section{Dataset Properties}

As a source for our dataset, we used the Stanford Dog dataset \cite{KhoslaYaoJayadevaprakashFeiFei_FGVC2011}, which contains 20,580 images, 120 breeds and bounding boxes for dogs. Since we are interested in bounding boxes for the dog's face rather than the entire body, we did not utilize the latter in the current study.

First, we selected a random subset with an equal amount of images per breed. Then, we filtered the images according to the following criteria: the image contains a single visible dog face, where the dog is in non-laboratory conditions (`in the wild'). Other dogs could be present, but their faces shouldn't be visible for unambiguity of detection. The resulting subset contains 7-40 images per breed (25 on average) of all 120 breeds (\dog images total), ranging in size from $100\times103$ to $1944\times2592$ pixels. Dogs in images have different sizes, colors, body and head poses, as well as different environments.

%\begin{figure}[h!]
%\centering
%\includegraphics[width=0.4\textwidth]{breeds_vertical.png}
%\caption{\textbf{} }\label{breeds}\vspace{-0.75em}
%\end{figure}

\begin{figure*}[t!]
\centering
\includegraphics[width=\textwidth]{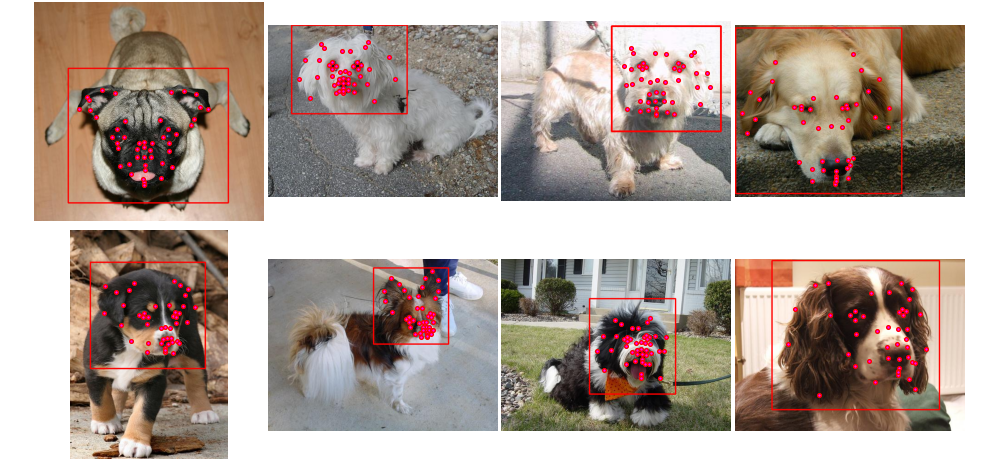}
\caption{\textbf{Examples of annotated images from the DogFLW.}}\label{ex}\vspace{-0.75em}
\end{figure*}

Each image is annotated with 46 facial landmarks using the CVAT platform\cite{CVAT2023} according to the scheme described above.  Unlike the CatFLW \cite{martvel2023catflw}, our dataset has occluded landmarks, all annotated landmarks have a visibility indicator ("0"~--- occluded, "1"~--- visible). Each image is also annotated with a  face bounding box, which encompasses the entire face of the animal, along with approximately 10\% of the surrounding space. This margin has proven crucial for training face detection models, as it prevents the cropping of important parts of the dog's face, such as the tips of the ears or the mouth. Figure \ref{ex} shows some examples of annotated images from the DogFLW dataset.

\section{Benchmarks}

For the evaluation, the dataset was randomly divided into the train (2,794 images) and test (480 images) sets. To provide correct metrics, we cropped all the faces by their detected bounding boxes in the preprocessing stage. All the models in this section were trained on a train set, and the final results are provided for the test set.

{\bf Metrics.} We use Normalized Mean Error ($NME_{iod}$) that preserves the relativity of the error regardless of the size of the image or the scale of the face on it. It is commonly utilized in landmark detection \cite{ZhangLLT14, cofw, cascade13}, and uses MAE as the basis and inter-ocular distance (distance between the outer corners of the two eyes, IOD) for normalization:

\begin{equation*}
NME_{iod} = \frac{1}{M \cdot N} \sum_{i=1}^{N} \sum_{j=1}^{M} \frac{\left\Vert {x_i}^j - {x'_i}^j \right\Vert_1}{iod_i},
\end{equation*}

\noindent where $M$ is the number of landmarks in the image, $N$ is the number of images in the dataset, ${x_i}^j$ and ${x'_i}^j$ --- the coordinates of the predicted and ground truth landmark respectively.

{\bf Experimental Setup.} We used the Ensemble Landmark Detector (ELD) \cite{martvel2024automated} model as a landmark detector, which is justified by its high performance on the CatFLW. We used the YOLOv8 \cite{Jocher_YOLO_by_Ultralytics_2023} model as a face detector and the EfficientNetV2S \cite{efnet} model as a backbone for the ensemble. All other parameters are similar to the original article.

The face detection model was trained using standard YOLOv8n training parameters for 100 epochs. The ensemble models were trained for 300 epochs using the mean squared error loss and the ADAM optimizer with a learning rate of $10^{-4}$ and a batch size of 16.

To train the models, we used the Google Colab cloud service with the NVIDIA TESLA V100 GPU.

{\bf Augmentation.} We randomly applied different augmentations (rotation, color balance adjustment, brightness and contrast modification, sharpness alteration, application of random blur masks, and addition of random noise) to the training data, doubling the size of the training set. Additionally, when training ensemble models, we mirrored symmetrical regions (ears and eyes) in order to further artificially increase the amount of training data.

\subsection{Dataset Size}

\begin{figure}[h!]
\includegraphics[width=0.47\textwidth]{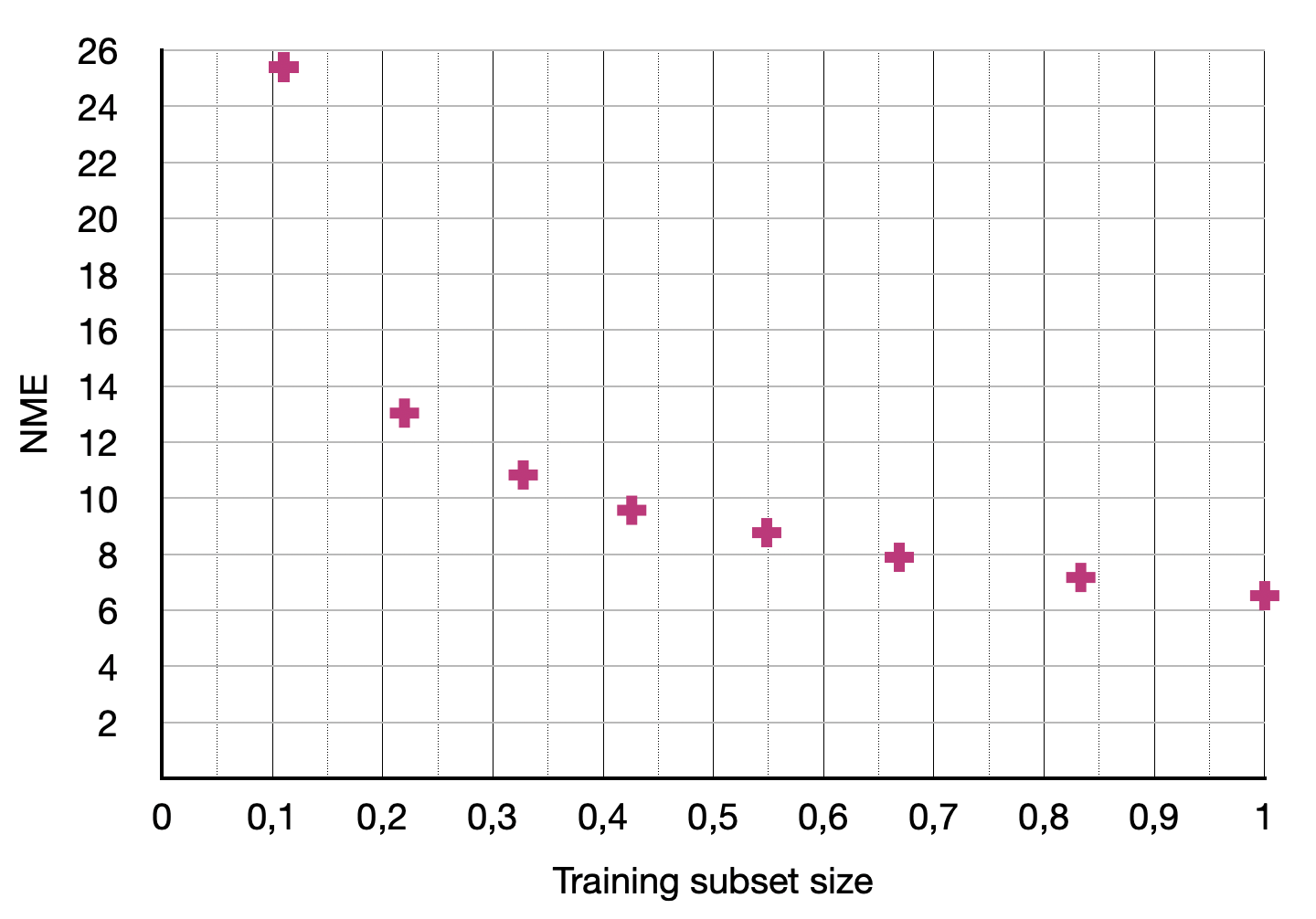}
\caption{\textbf{Impact of the training subset size on the normalised mean error.} The size of the subset is measured as a fraction of the total size of the training data.}\label{size}\vspace{-0.25em}
\end{figure}

\begin{figure*}[b!]
\centering
\includegraphics[width=\textwidth]{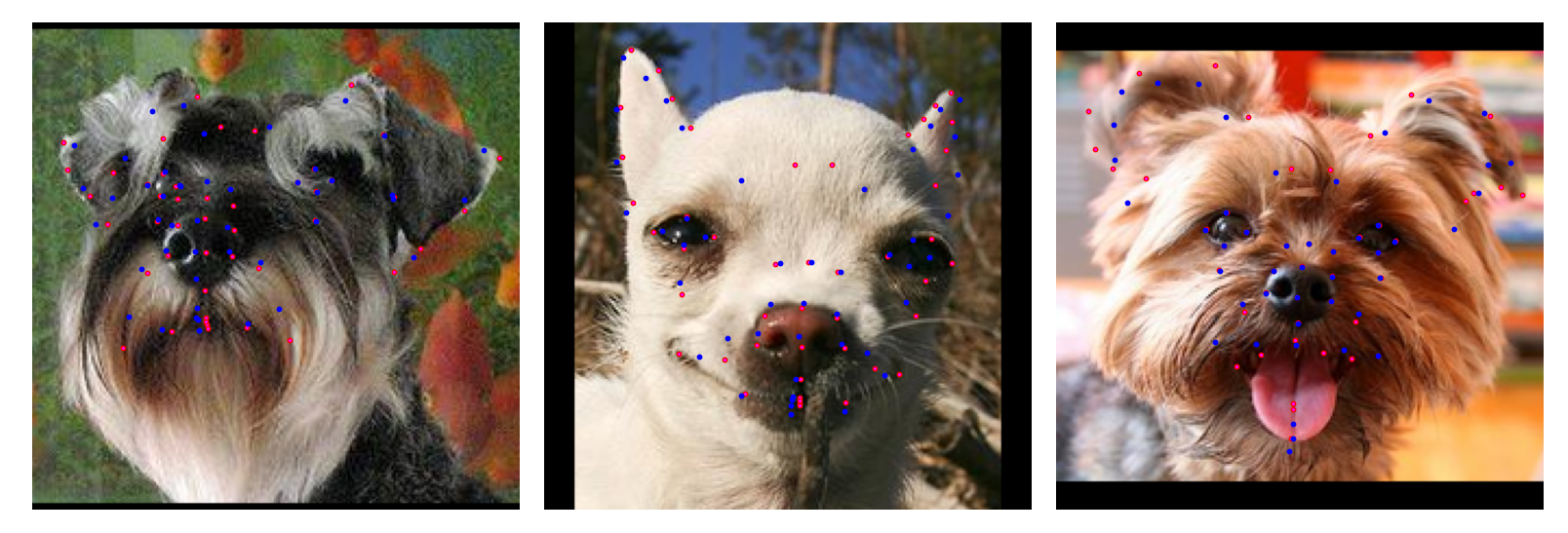}
\caption{\textbf{Landmark detection examples on pre-cropped images from the test set.} Red~--- ground truth, blue~--- predictions.}\label{det}\vspace{-0.75em}
\end{figure*}

In order to examine how the size of the training dataset affects the accuracy of landmark detection, we carried out a series of experiments, training the ELD on random subsets of the training data (averaging the results for 5 iterations) and gradually increasing their size. Figure \ref{size} indicates that the detection accuracy is approaching a plateau when the model is trained on 65\% of the data (1800 images). However, the accuracy ceiling has yet to be reached, indicating potential for further work into diversification and expanding the dataset size. The ELD model trained on the whole training set has a NME of 6.52. The detection examples are shown on Figure~\ref{det}.

\subsection{Ear Types}

While conducting experiments, it was observed that the accuracy of localizing landmarks in the ears is noticeably lower compared to other regions (Table \ref{tab:regions} shows the NME for different regions). This is presumably due to the diverse shapes and lengths of ears across different breeds, which may include previously unseen ear types and positions in the test set, particularly for floppy or half-floppy ear types.

\begin{table}[t!]
  \centering
  \begin{tabular}{@{}lccc@{}}
    \toprule
    Region & Eyes & Nose & Ears\\
    \midrule
    NME & 2.24 & 4.22 & 13.19 \\
    
    \bottomrule
  \end{tabular}
  \caption{Normalised mean error of landmark detection on the test set for different face regions}
  \label{tab:regions}
\end{table}

To test our hypothesis, we divided the data into three subsets: one with erect ears (pointy), another with hanging ears (floppy), and the third with other types (half-floppy). When dividing, we were guided by the average breed ear type in a relaxed state, obtaining a ratio of 31:50:19 for the training set and 29:52:19 for the test set.

Despite the dataset's predominance of breeds with floppy ear types, Table~\ref{tab:ears} shows that the accuracy of landmark detection on a test set for such dogs is less than for dogs with erect and half-floppy ears. This can be explained by the variety of lengths and positions of such ears.

\begin{table}[b!]
  \centering
  \begin{tabular}{@{}lccc@{}}
    \toprule
    Ear Type & \% in train set & NME total & NME ears only \\
    \midrule
    Pointy & 31 & 5.31 & 9.79 \\
    Floppy & 50 & 7.18 & 15.28 \\
    Half-floppy & 19 & 6.56 & 12.77 \\
    
    \bottomrule
  \end{tabular}
  \caption{Normalised mean error of landmark detection on the test set for dogs with different ear types}
  \label{tab:ears}
\end{table}

\subsection{Breeds}

Due to the uneven presence of breeds in the training set, the accuracy of landmark detection for rare breeds may be lower than for frequent breeds. However, it would be incorrect to say that only the number of samples affects the detection error value. The influence of the breed as an additional feature on the accuracy of detection is apparent --- many breeds have distinct fur length and texture, as well as facial anatomy. 

\begin{figure*}[!b]
%\hspace*{-7mm} % Adjust the value as needed to shift %the image to the left
\includegraphics[width=1\textwidth]{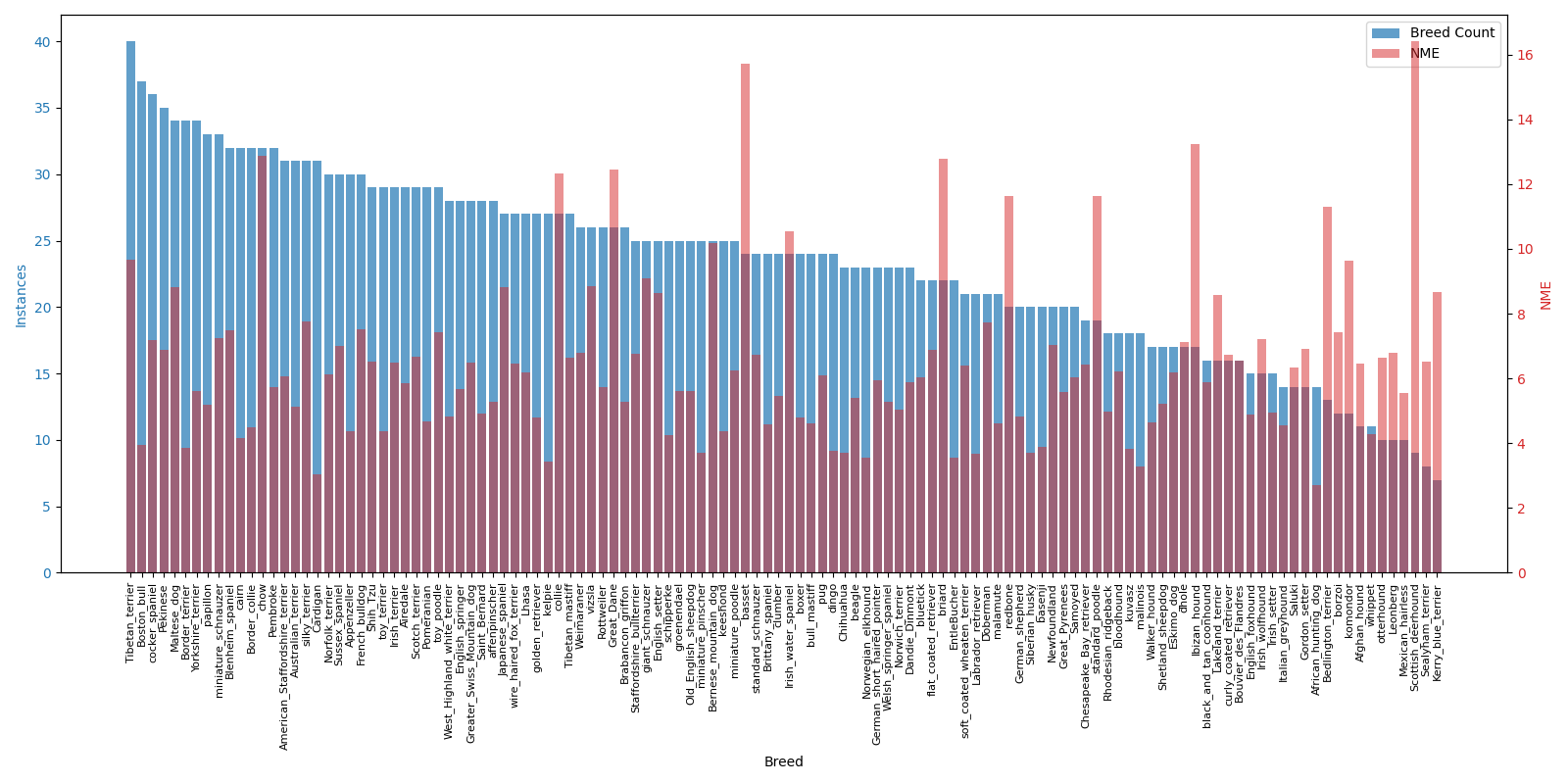}
\caption{\textbf{The distribution of images in the training set for each breed of dogs and the detection errors for the same breeds in the test set.}}\label{breeds}\vspace{-0.75em}
\end{figure*}

Figure \ref{breeds} shows the image distribution of dogs of each breed in the training set and the detection error for the same breeds in the test set. Based on the distribution, it can be observed that breeds with long fur covering facial features or extending on ears are the most difficult to detect accurately (\textit{Irish Water Spaniel, Briard, Standard Poodle, Bedlington Terrier, Scottish Deerhound, Komondor, Kerry Blue Terrier}). Moreover, in the Ear Types subsection, we demonstrated that detection has a high error on dogs with long, floppy ears, resulting in low accuracy in detecting facial landmarks for breeds with such ears (\textit{Basset, Redbone}). It is also worth noting that some breeds have non-obvious detection results. In some cases, this could be explained by a significant variation in the position of the ears (\textit{Ibizan Hound, Collie, Great Dane}) or they being obscure (\textit{Chow}), which can cause a significant detection error.

Breeds with short smooth facial fur (\textit{Cardigan, Kelpie, Miniature Pinscher, Dingo, Chihuahua, Malinois}, etc.), for which facial features are clearly distinguishable, have the lowest detection error on average.

\section{Conclusion}
The domestic dog is a highly social animal with a complex, elaborate communication system via facial signaling, which also underwent the most extreme morphological changes due to domestication and selective breeding. In this paper, we introduce a landmark scheme that is grounded in dog facial musculature. We further present an annotated dataset and benchmark detection results aimed at advancing the field of dog facial analysis. As expected, the model has the most difficulties with ear landmarks, which can be improved by enriching the dataset with more samples with diverse ear types. A similar strategy can be taken to improve the performance on certain breeds. Our future work includes utilizing the presented scheme and dataset to classify the internal states of dogs.

The presented dataset holds great promise for canine cognition, emotion, health, and welfare research. We hope that it will aid the scientific community in utilizing AI-driven methods to deepen our understanding of our best friends' behavior and emotional world.

%In this paper we present a CatFLW dataset of annotated cat faces. It includes \cat images of cats in various conditions with face bounding boxes and 48 facial landmarks for each cat. Each of the landmarks corresponds to a certain anatomical feature, which makes it possible to use these landmarks for various kinds of tasks on evaluating the internal state of cats. The proposed AI-assisted method of image annotation has significantly reduced the time for annotation of landmarks and reduced the amount of manual work during the preparation of the dataset. The created dataset opens up opportunities for the development of various computer vision models for detecting facial landmarks of cats and %, together with the proposed metrics for assessing the quality of such models,is can be the starting point for research in the field of behaviour observation and well-being of cats and animals in general.

\section*{Acknowledgements}

The research was supported by the Data Science Research
Center at the University of Haifa. We thank Ephantus Kanyugi for his contribution with data annotation and management. We thank Yaron Yossef for his technical support. 

%%%%%%%%% REFERENCES
{\small
\bibliographystyle{ieee_fullname}
\bibliography{egbib}
}

\end{document}